\documentclass[a4paper,12pt]{article}
\usepackage[utf8]{inputenc}
\usepackage[bitstream-charter]{mathdesign}
\usepackage{ragged2e}
\usepackage[english]{babel}
\usepackage{anysize}
\marginsize{2cm}{2cm}{2cm}{2cm}
\usepackage{amsmath}
\usepackage{hyperref}
\usepackage{textcomp,gensymb}
\usepackage{float}
\usepackage{lipsum}
\usepackage{graphicx}
\usepackage{caption}
\usepackage{subcaption}
\usepackage{url}
\usepackage{multicol}
\renewenvironment{abstract}
 {\par\noindent\textbf{\abstractname}\ \ignorespaces \\}
 {\par\noindent\medskip}
\usepackage{placeins}

\title{INFORME X}

\begin{document}

\begin{center}
\Large{A Distributed Privacy Preserving Model for the Detection of Alzheimer's Disease}
\vspace{0.4cm}

\normalsize
Paul K. Mandal
\vspace{0.1cm}

\textit{\small{The University of Texas at Austin}} \\
\href{mailto:mandal@utexas.edu}{\small\texttt{mandal@utexas.edu}}
\medskip

\normalsize

\end{center}
\line(1,0){482}
\begin{abstract}
\\
In the era of rapidly advancing medical technologies, the segmentation of medical data has become inevitable, necessitating the development of privacy preserving machine learning algorithms that can train on distributed data. Consolidating sensitive medical data is not always an option particularly due to the stringent privacy regulations imposed by the Health Insurance Portability and Accountability Act (HIPAA). In this paper, I introduce a HIPAA compliant framework that can train from distributed data. I then propose a multimodal vertical federated model for Alzheimer's Disease (AD) detection, a serious neurodegenerative condition that can cause dementia, severely impairing brain function and hindering simple tasks, especially without preventative care. This vertical federated learning (VFL) model offers a distributed architecture that enables collaborative learning across diverse sources of medical data while respecting privacy constraints imposed by HIPAA. The VFL architecture proposed herein offers a novel distributed architecture, enabling collaborative learning across diverse sources of medical data while respecting statutory privacy constraints. By leveraging multiple modalities of data, the robustness and accuracy of AD detection can be enhanced.  This model not only contributes to the advancement of federated learning techniques but also holds promise for overcoming the hurdles posed by data segmentation in medical research.
\end{abstract}
\line(1,0){482}
\medskip

\begin{multicols}{2}
\section{Introduction}
\label{sec:intro}

The increasing volume of medical data should be a priceless asset for advancing diagnostic machine learning (DML) models in healthcare \cite{binder2015big,huang2015promises,bates2014big}. Commonly, these large datasets are consolidated onto a centralized server for collaborators to investigate.  Unfortunately, consolidating medical data in such a fashion in order to train a centralized diagnostic machine learning model can be burdensome, costly, and impractical.
For example, the Health Insurance Portability and Accountability Act (HIPAA) \cite{act1996health,nosowsky2006health} and as updated through the Health Information Technology for Economic and Clinical Health Act (HITECH) \cite{hitech2009} mandate safeguards to protect any personal health information (PHI) from any breaches.  Additionally, consolidating medical data in order to  train a centralized model often can be an impractical endeavor.  The segmentation inherent in medical data across diverse sources imposes significant  limitations on both the depth and breadth of available information for robust DML training \cite{vest2010health}. Not surprisingly, consolidating and managing this data is fraught with administrative burdens, technical challenges, and considerable expenses.  Specifically, regional health information organizations (RHIOs) require US \$12 million each for development and incur \$2-3 million in operating costs \cite{scalise2006primer,terry2006rocky}.
    Just as a physician must grasp different types of clinical data (such as a history, physical examination, laboratory results, imaging studies, and questionnaires) to develop a unifying diagnosis, each data source used to train a DML model provides only a partial perspective.  DML demands novel methodologies capable of training from different modes of data while optimizing model development in any distributed environment. Thus, the utility of multimodal models becomes clearly apparent.  

Arguably, as explained above, prevailing approaches to multimodal model training practice data consolidation \cite{qiu2022multimodal} --fundamentally misaligned with the objectives of (1) preserving data segmentation, (2) protecting PHI, and (3) still optimizing model performance.  On the other hand, the integration of diverse data modalities holds promise for a more comprehensive understanding, particularly in addressing complex medical conditions.  Neuropsychiatric disorders, such as Alzheimer's Disease (AD), require interpreting a host of different objective and subjective data to establish a definitive diagnosis.  Because DML research on AD using multimodal data already has been reported, I chose to train a novel machine learning model-- specifically, a vertical federated learning model-- to detect AD.   The decision to address AD not only stems from a technical standpoint but also aligns with a broader, urgent societal need. Specifically, in 2024, an estimated 6.9 million or 1 in 9 (11\%) of all Americans aged 65 years and older are living with AD \cite{rajan2021population}. The Centers for Medicare and Medicaid Services shall pay out US\$ 206 billion for direct AD care \cite{skaria2022economic}.  Highlighting the overall the overall economic and associated burdens of AD to patients, their families, and society is the fact that there is an additionally estimated 18 billion unpaid hours for dementia-related care in 2022, valued at \$339.5 billion \cite{gaugler2022}. The number of Americans aged 65 and older not only will increase exponentially in the United States but also elsewhere-- as will the cost and other burdens of caring for AD.  Early detection and intervention of AD should improve the care and cut the cost of care for AD.  Therefore, there is a critical need for technical innovations facilitating early intervention and personalized care.

In this paper, a novel multimodal vertical federated learning (VFL)  framework that can be used for AD diagnosis is proposed and implemented.  First, an overview review of both horizontal and vertical federated learning initiatives is provided, along with previous research in DML in AD.  Next, the multimodal dataset, which consists of both demographic data and MRI images, is introduced. I then detail the experimental setup and model architectures, which can be readily reproduced by other investigators.  After discussing the results of this VFL for AD detection, I suggest how this framework can be improved and expanded so that it readily can be used in the clinical setting.

\section{Background}
\label{sec:Background}

\begin{figure*}[t]
    \centering
    \includegraphics[scale=2,width=0.9\textwidth]{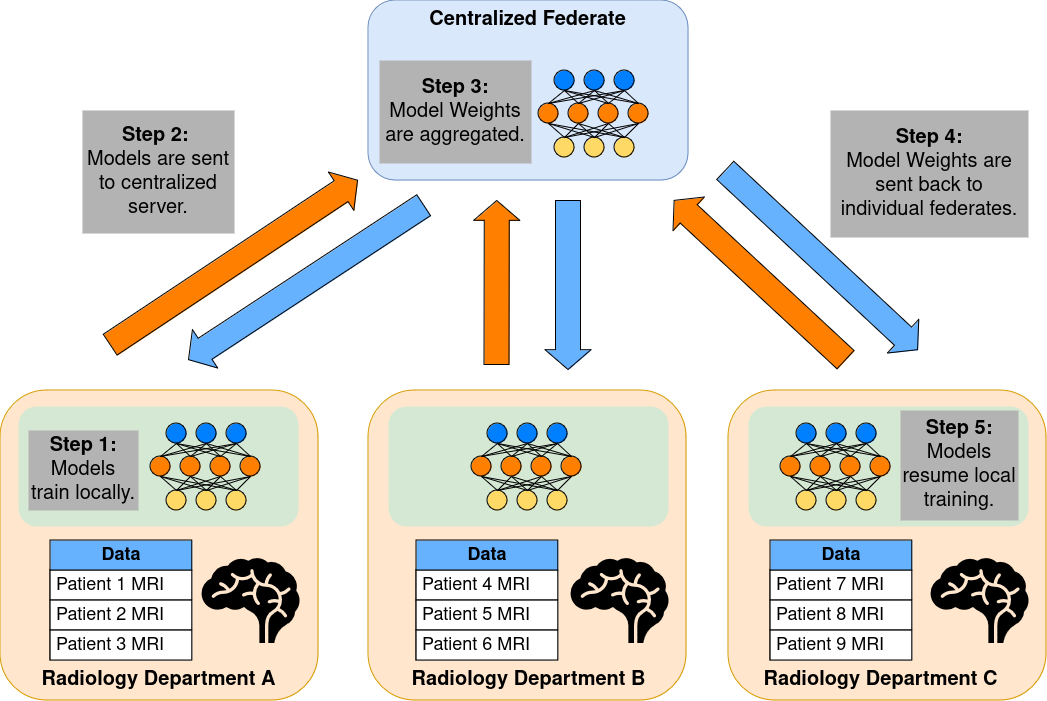}
    \caption{A Horizontal Federated Learning architecture for three radiology departments in different institutions to collaboratively train a single model using all of their data without medical record exchange.}
    \label{fig:hfl_arc}
\end{figure*}

\subsection{What Is Federated Learning?}
\label{sec:fml}
Federated learning is a decentralized machine learning framework which allows multiple disparate parties-- often on separate devices or servers-- to collaboratively train a global machine learning model without ever having to exchange their local data \cite{yang2019federated}. In scenarios wherever data security, patient privacy, and maintenance of data locality are necessary, federated learning can be a powerful tool. Rather than consolidating data in a central repository, federated learning leverages a federated architecture, where each data source (a "client" or "federate") maintains its own dataset and participates in model training through local computations. The global model then is updated iteratively, aggregating the model updates generated by each client while preserving the privacy and security of individual data sources.  Based on the framework architecture, federated learning models can be described as horizontal or vertical.

\subsection{Horizontal Federated Learning}

Horizontal Federated Learning (HFL) is defined as a scenario in which different parties share the same feature space but contain different data samples \cite{liu2022horizontal}.  After initialization of a global model, each client, possessing the same type of data, trains the model locally.  Clients collaborate to improve the global model by sharing insights from their locally collected data with the central server.  The central server then aggregates the models received from the clients, typically by averaging the model parameters. This averaging process ensures that the global model benefits from the diverse knowledge learned on different clients while preserving privacy.  One common method used in this process is Federated Averaging (FedAvg) \cite{yang2019federated}, where the central server orchestrates the federated learning process by specifically averaging the model updates and then redistributing the improved model to all clients.
Therefore, HFL is effective in scenarios where multiple federates are collecting similar data samples from multiple federates (or entities) without having to consolidate this data onto a centralized server.
    In a clinical setting, one could imagine a scenario where there are three hospitals. These hospitals all contain the same fields of data (such as magnetic resonance imaging or MRI). Radiology Department A at one hospital  may have MRIs for Patients 1-3 who were being evaluated for AD, Radiology Department B may MRIs of Patients 4-6, and Radiology Department C for Patients 7-9. In this case, an HFL model could be used for DML training since one mode of data (MRIs) from different patients are obtained from multiple federates in order to train a single neural network.  Local proprietary interests of the data and PHI protections thereby would be preserved since the actual images and patient identification is neither stored centrally nor shared among the collaborators .  The architecture of HFL is illustrated in figure \ref{fig:hfl_arc}.

\subsection{Clinical Applications for Horizontal Federated Learning}

There are many survey papers that both propose and implement HFL  in the healthcare research \cite{brauneckFedMedical,pfitznerFed}.
In terms of primary research of HFL models for AD detection training, Khalil and colleagues \cite{khalil2023federated} introduce a hardware accelerated horizontal federated multi-layer perceptron on field programmable gate arrays with 89\% accuracy on blood plasma data from the Alzheimer's Disease Neuroimaging Initiative (ADNI) \cite{ADNI}.  They also used K-nearest neighbors (KNN), support vector machines, decision trees, and logistic regression analyses.  

Similarly, MRIs from the ADNI and Australian Imaging, Biomarkers and Lifestyle (AIBL) \cite{AIBL} datasets were utilized to implement a federated domain adaptation framework via Transformer (FedDAvT) \cite{lei2023federated}.  This approach performed Unsupervised Deep Domain Adaptation and managed to achieve the following accuracy rates of: (1) 88.75\% for the AD vs. Normal Controls (NC), (2) 69.51\% for Mild Cognitive Impairment (MCI) vs. NC, and (3) 69.88\% for AD vs. MCI binary classification tasks.  

A  process called federated conditional mutual learning (FedCM)--which considers the similarity between clients-- improved these DML models that trained on MRI imaging \cite{huang2021federated}.  Using MRIs from the first version of Open Access Series of Imaging Studies (OASIS 1) \cite{marcus2007open}, ADNI, and AIBL datasets and then by extending upon FedAvg through FedCM, accuracy rates of  75.5\% accuracy rate on OASIS-1 and of 76.0\% on AIBL and ADNI were achieved. 

In each of the above studies of HFL models for AD detection, only one modality of data was used-- whether blood plasma protein levels or MRIs. Nevertheless, some investigators have attempted to train HFLs on multimodal data.  For example, Gao and colleagues performed a manifold projection of each electroencephalogram (EEG) signal, first by calculating a domain loss on each disparate signal and then by applying HFL to classify each signal \cite{gao2020hhhfl}. Although only one type of testing was analyzed (EEGs), manifold projection is necessary since each EEG signal comes from a different area of the brain and thus can be considered  multimodal data.   Moreover, this paper confirmed that all user data privacy was preserved.  Additionally, Peng and colleagues used federated graph convolution so that both imaging data and demographic data are converted into uniform high level features. A special graph generative adversarial network is then used to fill out missing nodes in the graph.  Therefore, the multimodal data is projected onto a uniform feature space, where FedAVG is applied to the locally trained graph for detecting both Alzheimer's Disease (AD) and Autism Spectrum Disorder (ASD) \cite{peng2022fedni}.  Both of the technologies described above succeeded in their objectives, training either a single federated neural network using an HFL process which would not perform well without the manifold projection \cite{gao2020hhhfl} or on multimodal data by using the same manifold projection \cite{peng2022fedni}. Unfortunately, projecting features onto a different latent space inevitably results in loss of information. Additionally, by losing the structure of the data, it is not possible to apply algorithms that best suit the data (e.g. Convolutional Neural Networks on images). Finally, the manifold projection of certain datasets is not possible because it is extremely difficult to project images down to the same latent space as demographic features, for example. 

\begin{figure*}[!t]
    \centering
    \includegraphics[scale=2,width=0.9\textwidth]{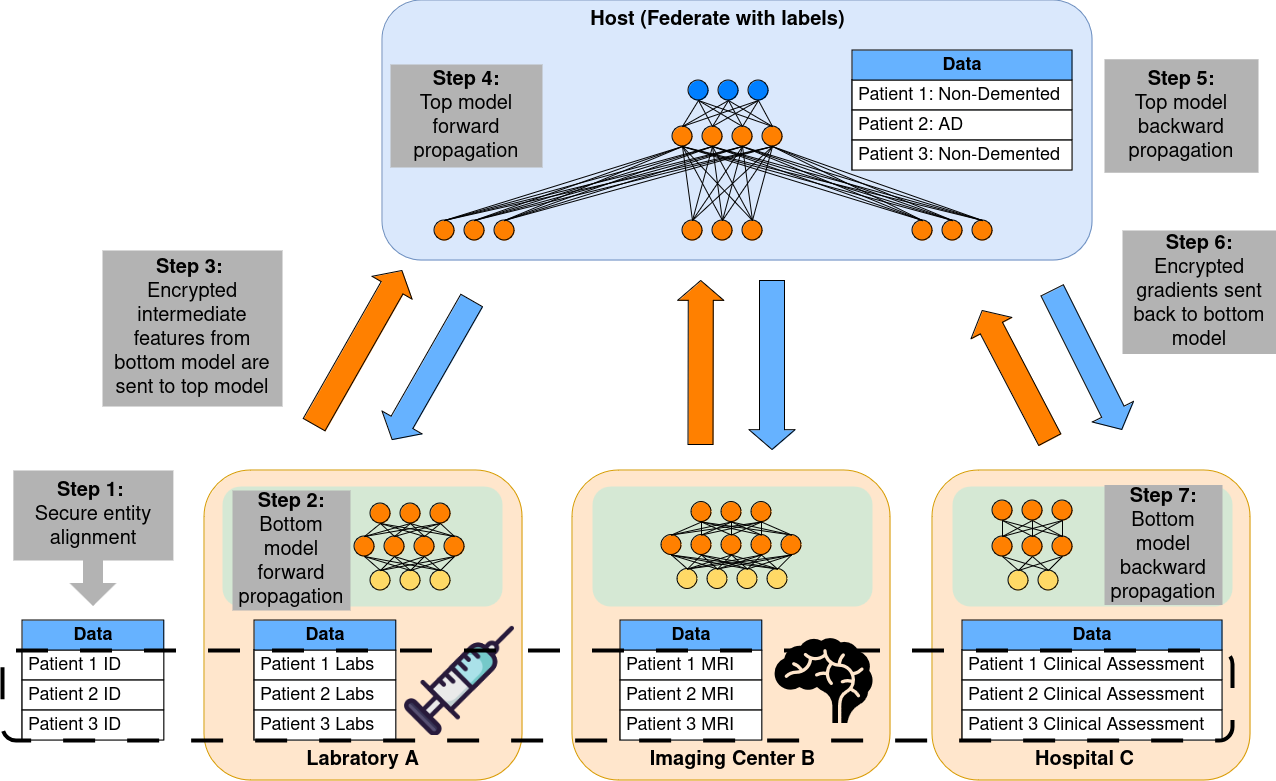}
    \caption{A Vertical Federated Learning Architecture where Labratory A has protein tests, Imaging Center B has MRI Data, and Hospital C possesses psychiatric assessments. Secure entity alignment occurs before training to ensure that the patients are "aligned" in order to perform the distributed training.}
    \label{fig:vfl_arc}
\end{figure*}

\subsection{Vertical Federated Learning}
Vertical Federated Learning (VFL)-- in contrast to HFL-- addresses instances in which multiple federates share the feature space and each training example is spread across different entities \cite{liu2022vertical}. Clients possess different features of a common entity, and the goal is to train a model that leverages all features for improved prediction and analysis. This approach enables predictive modeling and analytics across heterogeneous, multimodal data sources while preserving data privacy and security \cite{BlindFL}.

 In the clinical setting, for example, if a primary care physician (PCP) is concerned that a patient has early signs of AD, the PCP might order a test at Laboratory A and arrange for an MRI at Imaging Center B before referring the patient to a neurology colleague at Hospital C for a mental status assessment. VFL would allow for  multimodal training (or inference) of a global model at the level of a central server, utilizing of these different types of data obtained from a variety different institutions, without ever having to consolidate or centralize all the information in a single data warehouse. The architecture of vertical federated learning is shown in figure \ref{fig:vfl_arc}.

\begin{table*}[ht]
\centering
\caption{Features in the OASIS-2 Dataset}
\begin{tabular}{|p{0.15\linewidth} | p{0.7\linewidth}|}
\hline
\textbf{Feature} & \textbf{Description} \\ \hline
Group & Non-Demented, Demented, or Converted \\ \hline
Age &  	Age at time of image acquisition (years)  \\ \hline
Sex &  Sex (M or F) \\ \hline
EDUC & Years of education  \\ \hline
SSE & Socioeconomic status as assessed by the Hollingshead Index of Social Position \cite{hollingshead} \\ \hline
MMSE & Mini-Mental State Examination score \cite{minimentalstate}  \\ \hline
CDR & 	Clinical Dementia Rating \cite{morris1993clinical}  \\ \hline
ASF & Atlas scaling factor (unitless). Computed scaling factor that transforms native-space brain and skull to the atlas target (i.e., the determinant of the transform matrix) \cite{buckner2004unified} \\ \hline
eTIV & 	Estimated total intracranial volume (cm\textsuperscript{3}) \cite{buckner2004unified} \\ \hline
nWBV & Normalized whole-brain volume, expressed as a percent of all voxels in the atlas-masked image that are labeled as gray or white matter by the automated tissue segmentation process \cite{fotenos2005normative} \\ \hline
\end{tabular}
\label{tab:oasis}
\end{table*}

\subsection{Clinical Applications for Vertical Federated Learning}

 VFL for disease diagnosis is far less common.  Nevertheless, two notable publications are worth discussion. Firstly, Zhang and colleagues \cite{zhang2021aegis}  focused on the security of VFL systems using an example from the Federated Technology Enabler (FATE) \cite{liu2021fate}.  Developed by Webank's AI department, FATE remains  the world's first industrial-grade federated learning open source framework,  enabling enterprises and institutions to collaborate on data \FloatBarrier \noindent while protecting data security and privacy.  Ju et al. \cite{ju2020federated} adapted the manifold projection proposed in Gao et al. \cite{gao2020hhhfl} and applied it to federated transfer learning allowing for increased diagnostic accuracy by leveraging data diversity.

Based on the current literature review, both HFL and VFL would be ideal for DML in healthcare.  Since transfer of PHI data to a central data warehouse or among collaborators never has to occur, all federated learning models inherently would be HIPAA- and HITECH-compliant; dominion over data would remain locally.  The choice between and HFL and VFL model should be based on data distribution. If only a single type of data (such as MRI imaging) is going to be used for training a DML, then a HFL framework readily can be executed.  Although multimodal data training can occur with HFL, either (1) the multimodal features must be present on every federate or (2) the features must project down to the same latent space, inevitably leading to a loss of information and accuracy.  On the other hand, VFL could have the additional benefit of being able to handle diverse datasets.  While both the security and improved accuracy of a VFL model in healthcare have been established, to date, no one has investigated how a VFL model could be trained on multimodal data. 

\section{Methods}

\subsection{Dataset and Preprocessing of the Data}
\label{sec:data}
The second version of Open Access Series of Imaging Studies (OASIS-2) is comprised of a longitudinal cohort of 150 individuals aged 60 to 96 years \cite{oasis2}. All subjects were right-handed and consisted of both male and female subjects.  Table \ref{tab:oasis} summarizes the list of features provided in the OASIS-2 dataset.

The ASF, eTTV, and nWBV required MRI brain scans.  Each participant underwent two or more T1-weighted MRI scanning sessions, with a minimum one-year interval in between sessions, resulting in a total of 373 imaging sessions. There are 3 to 4 individual images used from each MRI scanning session. 

\begin{figure*}[ht]
    \centering
    \includegraphics[scale=2,width=0.7\textwidth]{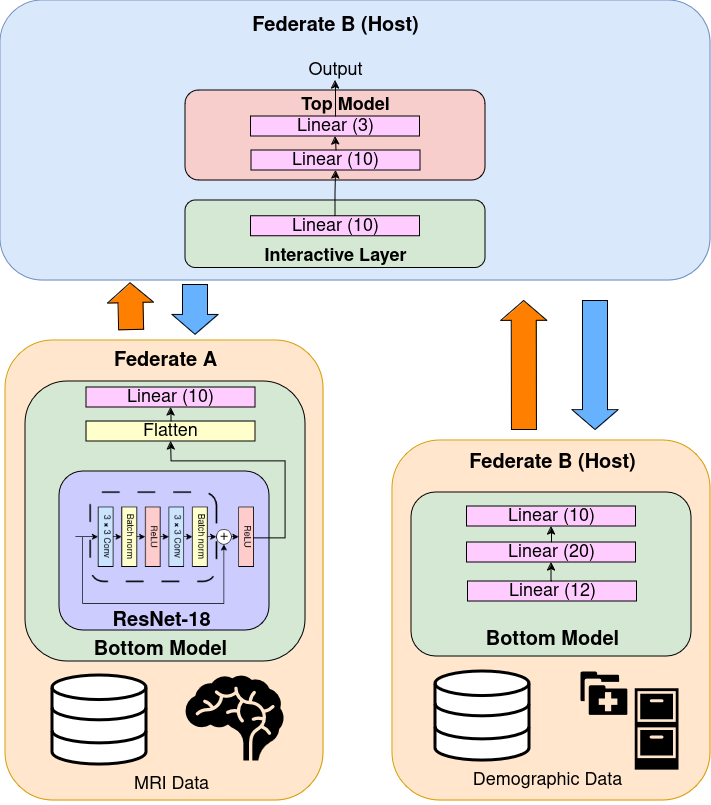}
    \caption{A Multimodal Vertical Federated Learning Model for the Diagnosis of Alzheimer's Dementia.}
    \label{fig:experiment}
\end{figure*}

\begin{figure}
    \centering
    \includegraphics[scale=2,width=0.7\textwidth]{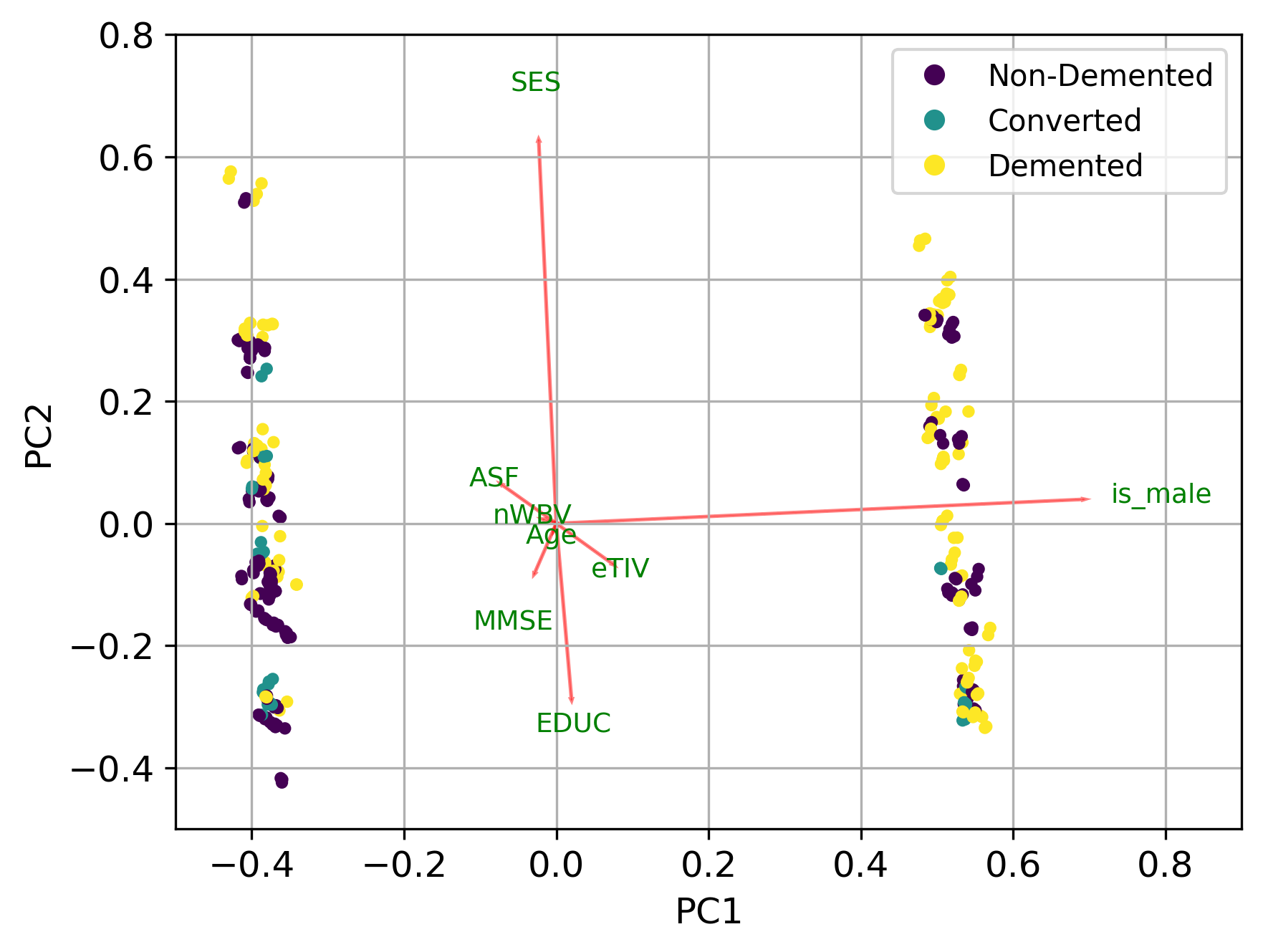}
    \caption{Scatter plot of the dataset projected onto two principal components with loading vectors}
    \label{fig:PCA_plot}
\end{figure}

Among the participants, 72 individuals \FloatBarrier \noindent maintained a non-demented status throughout the study. 64 individuals were initially classified as demented during their initial assessments and retained this classification in subsequent scans, including 51 individuals with mild to moderate Alzheimer’s disease. Most notably, 14 subjects initially characterized as non-demented were subsequently diagnosed with dementia. These subjects are labeled as "converted" on the study.  The scatterplot in figure \ref{fig:PCA_plot} illustrates the loading vectors of each demographic and clincal feature and each patient is projected down to a 2-dimensional space, using principal component analysis.

After excluding all patient records that had any missing features, the dataset comprised of 142 patients, 354 imaging sessions, and 1297 MRI images. I then normalized the values of the features from zero to one, choosing the "Group" feature as the label. The first 102 patients comprised the training set, 20 patients as the validation set, and another 20 patients as the test set. As mentioned earlier, each patient had multiple MRI images. Therefore, our training set included 952 records, the validation set had 181 records, and the test set consisted of 164 records. Since our model cannot train on patient-level data, results are reported at the record level.

\begin{figure*}[t]
    \centering
    \includegraphics[scale=2,width=0.9\textwidth]{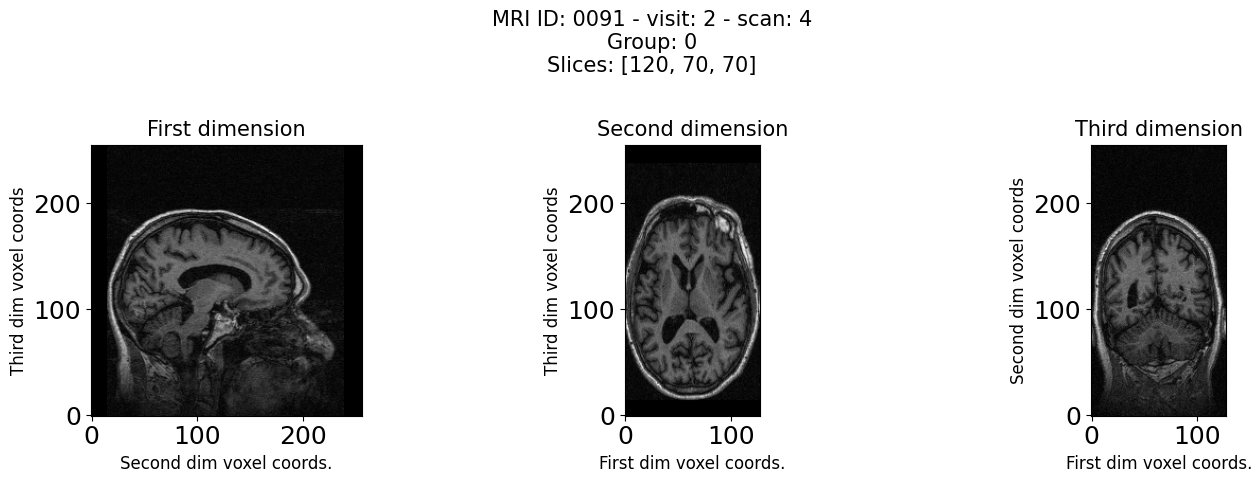}
    \caption{Cross sectional MRI Image}
    \label{fig:mri}
\end{figure*}

MRI images were imported in the form of NIfTI (Neuroimaging Informatics Technology Imitative) images \cite{cox2004sort} and then converted into tensors with NiBabel \cite{garyfallidis2014dipy}. Although a FATE compatible data loader to load the entire NiFTI images could have been used and writing a dedicated dataloader for NiFTI images would have been more useful in a clinical setting, such a feat would require approximately 1 GB of data per batch since our batch size is 32 and each NIfTI image is approximately 30 MB. Therefore, in the interest of efficiency, the MRI images were pre-processed into 256x256 JPEG format for training purposes. The 70th slice, 100th slice, and 125th slice were used to form the 3 channels of our JPEG image. Cross sectional views of a representative MRI image from our dataset is shown in figure \ref{fig:mri}.

\subsection{Experimental Model}
\label{sec:experiment}
Image data resided on Federate A. Demographic data along with the Clinical Dementia Rating resided on Federate B. On Federate A, the bottom model used a modified ResNet 18 architecture \cite{he2016deep}, a deep CNN employing residual connections to prevent vanishing gradients, architecture where the last layer of ResNet 18 was connected to a dense layer with 10 outputs. The accuracy of 3D CNNs compared to 2D CNNs on MRIs has been reported previouisly \cite{liang2021alzheimer}. Because-- on average-- 2D CNNs outperformed 3D counterparts \cite{liang2021alzheimer}, I opted to use a 2D CNN. On Federate B, the bottom model was a feed-forward neural network consisting of 12 inputs (each corresponding to a demographic feature), a hidden layer with 20 neurons, and an output layer with 10 neurons. These models were subsequently interconnected through an interactive layer with a size of 10 neurons. The top layer of this VFL model also was situated on Federate B because of the label's residence.  This top layer  consisted of a feed-forward neural network with 10 inputs and the resultant output. The Framework's architecture is illustrated in figure \ref{fig:experiment}.

FATE allows a user to submit a task such as a training or inference job through FATE Flow. For our purpose, we used the FATE Standalone client to simulate multiple federates on a single system (FATE will still simulate all of the network connections and processes necessary). A configuration for the architecture outlined in the previous paragraph was specified. The job then was submitted to FATE flow.

Unfortunately, FATE does not support loss metrics with a validation set. In order to get insights about overfitting and underfitting, a local version of this model was designed and trained in a similar manner described above, with the exception that the interactive layer was replaced with a linear layer and all models and data resided within the same system. Hence, propagation of gradients was done locally, rather than over the network. After designing the architectural framework, the model then was executed.

\section{Results}
\begin{figure}[H]
    \centering
    \includegraphics[scale=2,width=1\columnwidth]{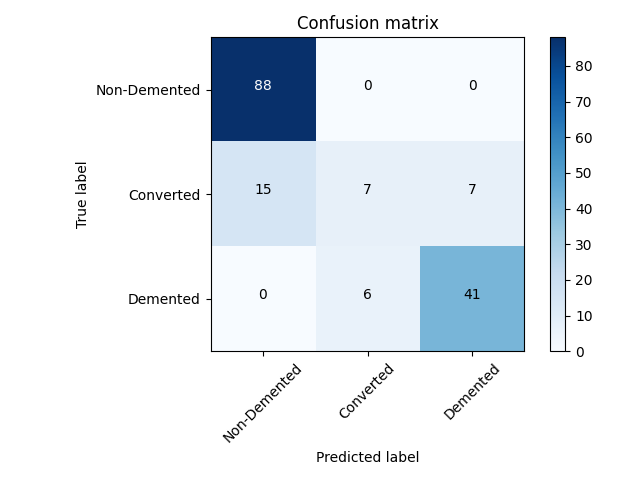}
    \caption{Confusion matrix of our proposed architecture for our 164 test records.}
    \label{fig:cm}
\end{figure}

This multimodal vertical federated neural network demonstrated promising results in the detection of Alzheimer's disease, achieving an overall accuracy rate of 82.9\% on the validation dataset. This performance metric demonstrates the model's ability to correctly classify instances of Alzheimer's, showcasing its potential utility in clinical applications. A 3×3 confusion matrix, containing True versus Predicted classifications for three classes of test subjects is provided in Figure \ref{fig:cm}. To contextualize this accuracy rate, this model's performance is compared with existing ligerature and benchmark models in Table \ref{tab:ad}.

\begin{table*}
\centering
\caption{Comparison of results for different federated methods for AD detection}
\begin{tabular}{|p{0.2\linewidth} | p{0.25\linewidth}|p{0.15\linewidth}|p{0.07\linewidth}|p{0.08\linewidth}|}
\hline
\textbf{Paper} & \textbf{Dataset} & \textbf{Method} & \textbf{Type} & \textbf{Acc} \\ \hline
Khalil et al. \cite{khalil2023federated} & ADNI Plasma \cite{ADNI} & MLP & HFL & \textbf{89\%} \\ \hline
 &  & SVM & HFL & 62\% \\ \hline
 &   & KNN & HFL & 67\%\\ \hline
 &  & DT & HFL & 59\%\\ \hline
 &  & LR & HFL & 59\%\\ \hline
Lei et al. \cite{lei2023federated} & ADNI \cite{ADNI}, AIBL \cite{AIBL} & FedDAvT & HFL & \textbf{89\%} \\ \hline
Huang et al. \cite{huang2021federated} & ADNI \cite{ADNI}, OASIS 1 \cite{marcus2007open}, AIBL \cite{AIBL} & FedCM & HFL & N/A \\ \hline
 & OASIS 1 \cite{marcus2007open} & FedCM & HFL & 74.5\% \\ \hline
 & AIBL \cite{AIBL} & FedCM & HFL & 76\% \\ \hline
Mandal (2024) & OASIS 2 \cite{oasis2} & ResNet + DNN & \textbf{VFL} & 82\%\\ \hline
\end{tabular}
\label{tab:ad}
\end{table*}

\begin{figure*}
    \centering
    \includegraphics[scale=2,width=.7\textwidth]{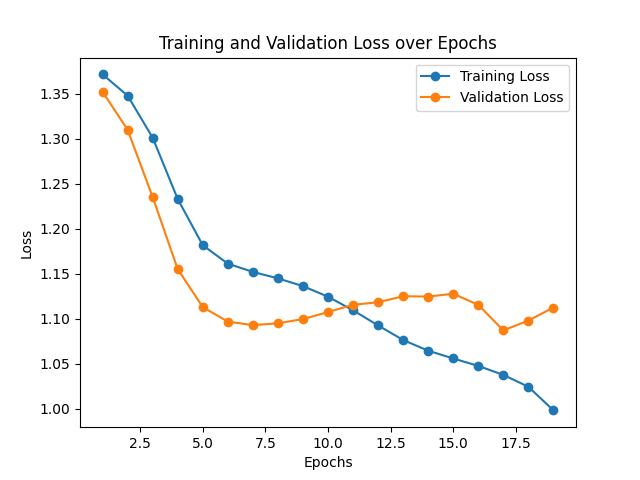}
    \caption{Training  and Validation Loss for the local VFL Model}
    \label{fig:loss}
\end{figure*}

Training and validation loss for the local version of this VFL architecture is illustrated in Figure \ref{fig:loss}. As explained above,  local training was used to tune the hyperparameters of this  federated model. With the chosen hyperparemeters, the loss plot indicates that the number of optimal epochs is 7. Training and validation loss for the local version of our architecture is illustrated in figure \ref{fig:loss}. 

\section{Discussion}
This paper proposed and then investigated how  a novel VFL platform for DML can be used for multimodal training.   As discussed above, multimodal training is paramount for DML when different types of data are used for clinical diagnoses.  Specific findings include the following:

Firstly, the VFL model could be trained for the diagnosis of AD, with an accuracy rate of 82.9\%.  This diagnostic accuracy rate is well within the results of other investigators who applied  HFL for AD detection and reported accuracy rates between 62\% and 89\% \cite{khalil2023federated, lei2023federated, huang2021federated}. Despite the accuracy rate obtained, one limitation of the study is OASIS-2 dataset's small sample size. The OASIS-2 dataset only had 150 patients, which made it challenging to train a robust neural network. Splitting the train, validation, and test sets on the imaging session or MRI image instead of on the patient axis may have yielded a higher accuracy rate.  However, such a strategy also could have resulted in an over-fit because of the small sample size.
    Moving forward, the OASIS-3 dataset \cite{oasis3} consists of 1,379 patients. Additionally, more data has been included-- such as both amyloid and metabolic positron emission tomography scans, neuropsychiatric testing, genetic testing, and other clinical data. Future studies, therefore, could use OASIS-3 for multimodal training of a VFL architecture for AD detection.  OASIS-3 training of a VFL archtitecture most likely would produce a much more robust model trained on more modalities of data.
    Although training on a wide array of different datasets-- as found in OASIS-3-- should increase diagnostic accuracy, the availability of more data in a clinical setting could bring about further challenges if used to train the herein described VFL model.  For example, the assumption  that all of the demographic features rely on one federate and the image data rely on another might neither be realistic or feasible. To obviate this hurdle, one option would be to apply the Cascade Vertical Federated Learning (CVFL) framework \cite{xia2021vertical} . CVFL  decomposes the neural network into subnetworks, thereby allowing the utilization of partitioned labels for privacy-preserving training. CVFL allows for VFL training when some of the labels are distributed horizontally. Such a framework is advantageous particularly in scenarios where features and labels are distributed among different parties,  Therefore, CVFL aligns with most clinical settings, in which data and labels often are distributed across multiple federates. The potential implementation of CVFL in future iterations of our VFL research promises a more robust and clinically relevant approach to federated learning within the healthcare domain.

Secondly, by using fully homomorphic encryption to prevent the leakage of model weights, my VFL model is privacy-preserving.  Since federated learning does not require sending personal health information to the central server, it inherently is HIPAA compliant.   The importance of security and privacy protections is a costly concern. With regard to preserving personal health information, in 2019, it was reported that the total cost of healthcare sector data breaches totaled \$8 million dollars, or \$400 dollars per each medical record.  In the first 6 months of 2024, there already have been 340 data breaches, affecting 42,250,702 individuals \cite{hhs_breach_portal,hhs_cost_analysis}.  By the end of this year, a total cost of 11 million dollars or 500 dollars per medical record could be anticipated.  Thus, DML must have safeguard around PHI, and-- as discussed above--  all federated learning frameworks inherently would protect PHI.

Thirdly, this paper demonstrates how current, open-source federated libraries (i.e. FATE) and deep learning algorithms (i.e. ResNet-18) readily can be applied to formulate a VFL architecture for clinical diagnoses.  Based on the above findings, either VFL or CVFL should be the primary focus whenever multimodal training is required in DML.  These models not only work on diverse datasets simultaneously but also provide improved accuracy rates and PHI protection

In spite of the challenges and limitations outlined above, the work herein holds significant promise for further investigation of VFL implementation in DML. As reviewed above, a notable gap in the literature-- regarding the use of VFL to address data segmentation challenges that arise when different features and imagery for a patient are distributed across multiple parties-- currently exists.  By determining whether a VFL framework for AD detection could be implemented, the challenges of data segmentation along the feature axis was successfully addressed while still protecting personal health information. Future VFL modeling frameworks should lead to further innovations in applying machine learning to clinical medicine.

\section*{Acknowledgements}
The author of this paper would like to thank Aloke K. Mandal, M.D., Ph.D. for his clinical input and assistance in editing this manuscript.

Data was provided by OASIS: Longitudinal: Principal Investigators: D. Marcus, R, Buckner, J. Csernansky, J. Morris; P50 AG05681, P01 AG03991, P01 AG026276, R01 AG021910, P20 MH071616, U24 RR021382.

\section*{Published Version}
The citation for the published version of this paper is: \\
Mandal, P.K. A distributed privacy preserving model for the detection of Alzheimer’s disease. \textit{Neural Computing and Applications} (2024). \href{https://doi.org/10.1007/s00521-024-10419-4}{\texttt{https://doi.org/10.1007/s00521-024-10419-4}}.

The published copy of this paper may be viewed for free at \href{https://rdcu.be/dU8yX}{\texttt{https://rdcu.be/dU8yX}}. The differences between this version and the published version are mainly minor formatting issues.

\medskip
\bibliography{mybibliography} 

\begin{thebibliography}{10}

\bibitem{binder2015big}
Harald Binder and Maria Blettner.
\newblock Big data in medical science—a biostatistical view: Part 21 of a series on evaluation of scientific publications.
\newblock {\em Deutsches {\"A}rzteblatt International}, 112(9):137, 2015.

\bibitem{huang2015promises}
Tao Huang, Liang Lan, Xuexian Fang, Peng An, Junxia Min, and Fudi Wang.
\newblock Promises and challenges of big data computing in health sciences.
\newblock {\em Big Data Research}, 2(1):2--11, 2015.

\bibitem{bates2014big}
David~W Bates, Suchi Saria, Lucila Ohno-Machado, Anand Shah, and Gabriel Escobar.
\newblock Big data in health care: using analytics to identify and manage high-risk and high-cost patients.
\newblock {\em Health affairs}, 33(7):1123--1131, 2014.

\bibitem{act1996health}
{United States Public Law 104-191}.
\newblock Health insurance portability and accountability act of, 1996.
\newblock 110 Stat. 1936; August 21, 1996. Available from: \url{https://www.govinfo.gov/content/pkg/PLAW-104publ191/pdf/PLAW-104publ191.pdf}; Accessed on July 15, 2024.

\bibitem{nosowsky2006health}
Rachel Nosowsky and Thomas~J Giordano.
\newblock The health insurance portability and accountability act of 1996 (hipaa) privacy rule: implications for clinical research.
\newblock {\em Annu. Rev. Med.}, 57:575--590, 2006.

\bibitem{hitech2009}
{United States Public Law 111-5}.
\newblock Title xiii: Health information technology for economic and clinical health act of, 2009.
\newblock 123 Stat. 227; February 17, 2009. Available from: \url{https://www.govinfo.gov/content/pkg/PLAW-111publ5/pdf/PLAW-111publ5.pdf}; Accessed on July 15, 2024.

\bibitem{vest2010health}
Joshua~R Vest and Larry~D Gamm.
\newblock Health information exchange: persistent challenges and new strategies.
\newblock {\em Journal of the American Medical Informatics Association}, 17(3):288--294, 2010.

\bibitem{scalise2006primer}
DA~Scalise.
\newblock Primer for building rhios.
\newblock {\em Hosp Health Netw}, 80:49--53, 2006.

\bibitem{terry2006rocky}
Ken Terry.
\newblock The rocky road to rhios.
\newblock {\em Medical economics}, 83(4):TCP8--TCP12, 2006.

\bibitem{qiu2022multimodal}
Shangran Qiu, Matthew~I Miller, Prajakta~S Joshi, Joyce~C Lee, Chonghua Xue, Yunruo Ni, Yuwei Wang, Ileana De~Anda-Duran, Phillip~H Hwang, Justin~A Cramer, et~al.
\newblock Multimodal deep learning for alzheimer’s disease dementia assessment.
\newblock {\em Nature communications}, 13(1):3404, 2022.

\bibitem{rajan2021population}
Kumar~B Rajan, Jennifer Weuve, Lisa~L Barnes, Elizabeth~A McAninch, Robert~S Wilson, and Denis~A Evans.
\newblock Population estimate of people with clinical alzheimer's disease and mild cognitive impairment in the united states (2020--2060).
\newblock {\em Alzheimer's \& dementia}, 17(12):1966--1975, 2021.

\bibitem{skaria2022economic}
Anita~Pothen Skaria.
\newblock The economic and societal burden of alzheimer disease: managed care considerations.
\newblock {\em The American journal of managed care}, 28(10 Suppl):S188--S196, 2022.

\bibitem{gaugler2022}
Joseph Gaugler, Bryan James, Tricia Johnson, Jessica Reimer, Michele Solis, Jennifer Weuve, Rachel~F Buckley, and Timothy~J Hohman.
\newblock 2022 alzheimer's disease facts and figures.
\newblock {\em Alzheimers \& Dementia}, 18(4):700--789, 2022.

\bibitem{yang2019federated}
Qiang Yang, Yang Liu, Tianjian Chen, and Yongxin Tong.
\newblock Federated machine learning: Concept and applications, 2019.

\bibitem{liu2022horizontal}
Dianqi Liu, Liang Bai, Tianyuan Yu, and Aiming Zhang.
\newblock Towards method of horizontal federated learning: A survey.
\newblock In {\em 2022 8th International Conference on Big Data and Information Analytics (BigDIA)}, pages 259--266, 2022.

\bibitem{brauneckFedMedical}
Alissa Brauneck, Louisa Schmalhorst, Mohammad~Mahdi Kazemi~Majdabadi, Mohammad Bakhtiari, Uwe V{\"o}lker, Jan Baumbach, Linda Baumbach, and Gabriele Buchholtz.
\newblock Federated machine learning, privacy-enhancing technologies, and data protection laws in medical research: Scoping review, Mar 2023.

\bibitem{pfitznerFed}
Bjarne Pfitzner, Nico Steckhan, and Bert Arnrich.
\newblock Federated learning in a medical context: A systematic literature review, jun 2021.

\bibitem{khalil2023federated}
Kasem Khalil, Mohammad Mahbubur~Rahman Khan~Mamun, Ahmed Sherif, Mohamed~Said Elsersy, Ahmad Abdel-Aliem Imam, Mohamed Mahmoud, and Maazen Alsabaan.
\newblock A federated learning model based on hardware acceleration for the early detection of alzheimer’s disease.
\newblock {\em Sensors}, 23(19):8272, 2023.

\bibitem{ADNI}
Susanne~G. Mueller, Michael~W. Weiner, Leon~J. Thal, Ronald~C. Petersen, Clifford~R. Jack~Jr., William Jagust, John~Q. Trojanowski, Arthur~W. Toga, and Laurel Beckett.
\newblock The alzheimer's disease neuroimaging initiative.
\newblock {\em NeuroImage}, 26(3):1011--1012, 2005.

\bibitem{AIBL}
Kathryn~A. Ellis, Ashley~I. Bush, David Darby, Daniela De~Fazio, Jonathan Foster, Peter Hudson, Nicola~T. Lautenschlager, Nat Lenzo, Ralph~N. Martins, Paul Maruff, Colin~L. Masters, Ashley Milner, Kerryn~E. Pike, Christopher Rowe, Greg Savage, Cassandra Szoeke, Kevin Taddei, and Victor~L. Villemagne.
\newblock The australian imaging, biomarkers and lifestyle (aibl) study of aging: methodology and baseline characteristics of 1112 individuals recruited for a longitudinal study of alzheimer's disease.
\newblock {\em International Psychogeriatrics}, 21(4):672--687, 2009.

\bibitem{lei2023federated}
Baiying Lei, Yun Zhu, Enmin Liang, Peng Yang, Shaobin Chen, Huoyou Hu, Haoran Xie, Ziyi Wei, Fei Hao, Xuegang Song, et~al.
\newblock Federated domain adaptation via transformer for multi-site alzheimer’s disease diagnosis.
\newblock {\em IEEE Transactions on Medical Imaging}, 2023.

\bibitem{huang2021federated}
Ya-Lin Huang, Hao-Chun Yang, and Chi-Chun Lee.
\newblock Federated learning via conditional mutual learning for alzheimer’s disease classification on t1w mri.
\newblock In {\em 2021 43rd Annual International Conference of the IEEE Engineering in Medicine \& Biology Society (EMBC)}, pages 2427--2432. IEEE, 2021.

\bibitem{marcus2007open}
Daniel~S Marcus, Tracy~H Wang, Jamie Parker, John~G Csernansky, John~C Morris, and Randy~L Buckner.
\newblock Open access series of imaging studies (oasis): cross-sectional mri data in young, middle aged, nondemented, and demented older adults.
\newblock {\em Journal of cognitive neuroscience}, 19(9):1498--1507, 2007.

\bibitem{gao2020hhhfl}
Dashan Gao, Ce~Ju, Xiguang Wei, Yang Liu, Tianjian Chen, and Qiang Yang.
\newblock Hhhfl: Hierarchical heterogeneous horizontal federated learning for electroencephalography, 2020.

\bibitem{peng2022fedni}
Liang Peng, Nan Wang, Nicha Dvornek, Xiaofeng Zhu, and Xiaoxiao Li.
\newblock Fedni: Federated graph learning with network inpainting for population-based disease prediction.
\newblock {\em IEEE Transactions on Medical Imaging}, 2022.

\bibitem{liu2022vertical}
Yang Liu, Yan Kang, Tianyuan Zou, Yanhong Pu, Yuanqin He, Xiaozhou Ye, Ye~Ouyang, Ya-Qin Zhang, and Qiang Yang.
\newblock Vertical federated learning, 2022.

\bibitem{BlindFL}
Fangcheng Fu, Huanran Xue, Yong Cheng, Yangyu Tao, and Bin Cui.
\newblock Blindfl: Vertical federated machine learning without peeking into your data, 2022.

\bibitem{hollingshead}
August Hollingshead.
\newblock Two-factor index of social position.
\newblock {\em Yale University Press}, 1965.

\bibitem{minimentalstate}
Marshal~F Folstein, Susan~E Folstein, and Paul~R McHugh.
\newblock “mini-mental state”: a practical method for grading the cognitive state of patients for the clinician.
\newblock {\em Journal of psychiatric research}, 12(3):189--198, 1975.

\bibitem{morris1993clinical}
John~C Morris.
\newblock The clinical dementia rating (cdr): current version and scoring rules.
\newblock {\em Neurology}, 1993.

\bibitem{buckner2004unified}
Randy~L Buckner, Denise Head, Jamie Parker, Anthony~F Fotenos, Daniel Marcus, John~C Morris, and Abraham~Z Snyder.
\newblock A unified approach for morphometric and functional data analysis in young, old, and demented adults using automated atlas-based head size normalization: reliability and validation against manual measurement of total intracranial volume.
\newblock {\em Neuroimage}, 23(2):724--738, 2004.

\bibitem{fotenos2005normative}
Anthony~F Fotenos, AZ~Snyder, LE~Girton, JC~Morris, and RL~Buckner.
\newblock Normative estimates of cross-sectional and longitudinal brain volume decline in aging and ad.
\newblock {\em Neurology}, 64(6):1032--1039, 2005.

\bibitem{zhang2021aegis}
Cengguang Zhang, Junxue Zhang, Di~Chai, and Kai Chen.
\newblock Aegis: A trusted, automatic and accurate verification framework for vertical federated learning, 2021.

\bibitem{liu2021fate}
Yang Liu, Tao Fan, Tianjian Chen, Qian Xu, and Qiang Yang.
\newblock Fate: An industrial grade platform for collaborative learning with data protection.
\newblock {\em Journal of Machine Learning Research}, 22(226):1--6, 2021.

\bibitem{ju2020federated}
Ce~Ju, Dashan Gao, Ravikiran Mane, Ben Tan, Yang Liu, and Cuntai Guan.
\newblock Federated transfer learning for eeg signal classification.
\newblock In {\em 2020 42nd annual international conference of the IEEE engineering in medicine \& biology society (EMBC)}, pages 3040--3045. IEEE, 2020.

\bibitem{oasis2}
Daniel~S Marcus, Anthony~F Fotenos, John~G Csernansky, John~C Morris, and Randy~L Buckner.
\newblock Open access series of imaging studies: longitudinal mri data in nondemented and demented older adults.
\newblock {\em Journal of cognitive neuroscience}, 22(12):2677--2684, 2010.

\bibitem{cox2004sort}
Robert~W Cox, John Ashburner, Hester Breman, Kate Fissell, Christian Haselgrove, Colin~J Holmes, Jack~L Lancaster, David~E Rex, Stephen~M Smith, Jeffrey~B Woodward, et~al.
\newblock A (sort of) new image data format standard: Nifti-1.
\newblock In {\em 10th Annual Meeting of the Organization for Human Brain Mapping}, volume~22, page~01, 2004.

\bibitem{garyfallidis2014dipy}
Eleftherios Garyfallidis, Matthew Brett, Bagrat Amirbekian, Ariel Rokem, Stefan Van Der~Walt, Maxime Descoteaux, Ian Nimmo-Smith, and Dipy Contributors.
\newblock Dipy, a library for the analysis of diffusion mri data.
\newblock {\em Frontiers in neuroinformatics}, 8:8, 2014.

\bibitem{he2016deep}
Kaiming He, Xiangyu Zhang, Shaoqing Ren, and Jian Sun.
\newblock Deep residual learning for image recognition.
\newblock In {\em Proceedings of the IEEE conference on computer vision and pattern recognition}, pages 770--778, 2016.

\bibitem{liang2021alzheimer}
Gongbo Liang, Xin Xing, Liangliang Liu, Yu~Zhang, Qi~Ying, Ai-Ling Lin, and Nathan Jacobs.
\newblock Alzheimer’s disease classification using 2d convolutional neural networks.
\newblock In {\em 2021 43rd Annual International Conference of the IEEE Engineering in Medicine \& Biology Society (EMBC)}, pages 3008--3012. IEEE, 2021.

\bibitem{oasis3}
Pamela~J LaMontagne, Tammie~LS Benzinger, John~C Morris, Sarah Keefe, Russ Hornbeck, Chengjie Xiong, Elizabeth Grant, Jason Hassenstab, Krista Moulder, Andrei~G Vlassenko, et~al.
\newblock Oasis-3: longitudinal neuroimaging, clinical, and cognitive dataset for normal aging and alzheimer disease.
\newblock {\em MedRxiv}, pages 2019--12, 2019.

\bibitem{xia2021vertical}
Wensheng Xia, Ying Li, Lan Zhang, Zhonghai Wu, and Xiaoyong Yuan.
\newblock A vertical federated learning framework for horizontally partitioned labels, 2021.

\bibitem{hhs_breach_portal}
{Office for Civil Rights, U.S. Department of Health and Human Services}.
\newblock Breach portal: Notice to the secretary of hhs breach of unsecured protected health information.
\newblock \url{https://ocrportal.hhs.gov/ocr/breach/breach_report.jsf;jsessionid=5D3526C953D0AD18B1994D7C432D8A20}.
\newblock Accessed on July 15, 2024.

\bibitem{hhs_cost_analysis}
{Health Sector Cybersecurity Coordination Center, U.S. Department of Health and Human Services}.
\newblock A cost analysis of healthcare sector data breaches.
\newblock \url{https://www.hhs.gov/sites/default/files/cost-analysis-of-healthcare-sector-data-breaches.pdf}, April 12 2019.
\newblock Accessed on July 15, 2024.

\end{thebibliography}
\newpage

\end{multicols}
\end{document}